\def\year{2022}\relax
\documentclass[letterpaper]{article} 
\usepackage{aaai22}  
\usepackage{times}  
\usepackage{helvet}  
\usepackage{courier}  
\usepackage[hyphens]{url}  
\usepackage{graphicx} 
\usepackage{natbib}  
\usepackage{caption} 
\DeclareCaptionStyle{ruled}{labelfont=normalfont,labelsep=colon,strut=off} 
\usepackage{algorithm}
\usepackage{algorithmic}
\usepackage{booktabs}
\usepackage{multirow}
\usepackage{amsmath}
\usepackage{amssymb}
\usepackage[colorlinks,linkcolor=red]{hyperref}
\usepackage{newfloat}
\usepackage{listings}
\floatstyle{ruled}
\newfloat{listing}{tb}{lst}{}
\floatname{listing}{Listing}
\title{DCAN: Improving Temporal Action Detection via Dual Context Aggregation}
\author{
    Guo Chen\equalcontrib \quad
    Yin-Dong Zheng\equalcontrib \quad
    Limin Wang \quad
    Tong Lu\thanks{Corresponding author.}
}
\affiliations{
 State Key Lab for Novel Software Technology, Nanjing University, China\\
    \{chenguo1177, ydzheng0331\}@gmail.com, \{lmwang, lutong\}@nju.edu.cn
   
}

\usepackage{bibentry}

\begin{document}

\maketitle

\begin{abstract}
Temporal action detection aims to locate the boundaries of action in the video. 
The current method based on boundary matching enumerates and calculates all possible boundary matchings to generate proposals.
However, these methods neglect the long-range context aggregation in boundary prediction.
At the same time, due to the similar semantics of adjacent matchings, local semantic aggregation of densely-generated matchings cannot improve semantic richness and discrimination.
In this paper, we propose the end-to-end proposal generation method named {\em Dual Context Aggregation Network} (DCAN) to aggregate context on two levels, namely, boundary level and proposal level, for generating high-quality action proposals, thereby improving the performance of temporal action detection.
Specifically, we design the Multi-Path Temporal Context Aggregation (MTCA) to achieve smooth context aggregation on boundary level and precise evaluation of boundaries. 
For matching evaluation, Coarse-to-fine Matching (CFM) is designed to aggregate context on the proposal level and refine the matching map from coarse to fine. 
We conduct extensive experiments on ActivityNet v1.3 and THUMOS-14.
DCAN obtains an average mAP of 35.39\% on ActivityNet v1.3 and reaches mAP 54.14\% at IoU@0.5 on THUMOS-14, which demonstrates DCAN can generate high-quality proposals and achieve state-of-the-art performance.
We release the code at \url{https://github.com/cg1177/DCAN}.

\end{abstract}

\section{Indroduction}

With the explosive growth of Internet videos, video content's understanding and analysis technology have attracted more academia and industry attention.
Temporal action detection is to locate the boundaries of action instances and recognize action categories in untrimmed videos.
Video data is a stack of image frames, and the semantic changes between frames are more complicated to capture than the semantic changes between image pixels.
Therefore, compared with object detection, temporal action detection focuses more on processing and capturing the temporal information of the video. 

\begin{figure}[!ht]
\centering
\includegraphics[width=8.3cm]{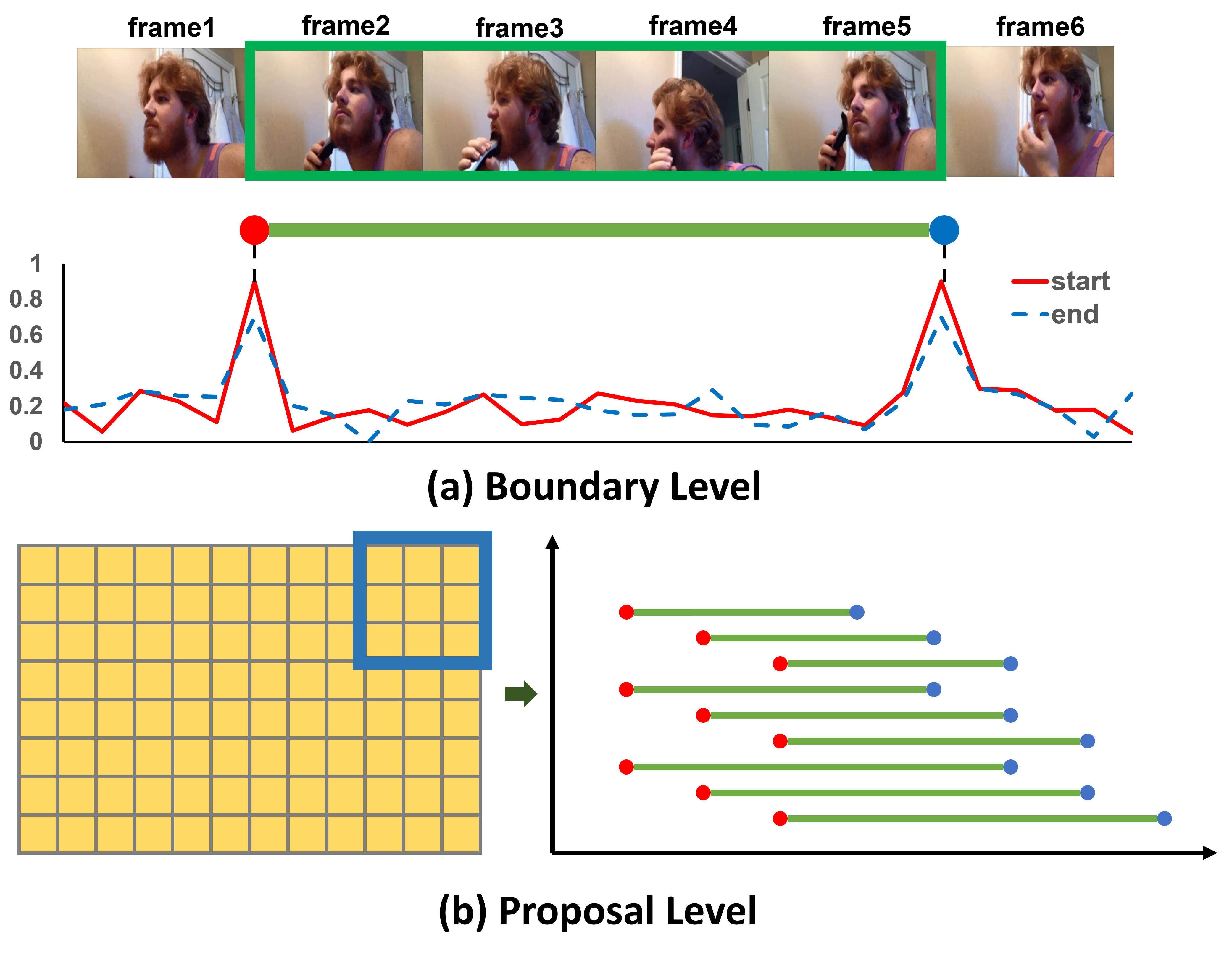}
\caption{Examples to illustrate the importance and difficulty of context aggregation on two levels. (a) The $\text{frame}_2$ and $\text{frame}_5$ with similar local context are difficult to be distinguished, which will generate many positive false proposals. (b) The temporal intervals of adjacent $3 \times 3$ matchings on the matching map are highly overlapped, lacking semantic discrimination and richness, and it is not easy to obtain effective semantic supplements by simple aggregation.}

\label{fig:intr}
\end{figure}

Similar to object detection, the temporal action detection methods can be divided into one-stage and two-stage methods. The one-stage methods simultaneously locate the action boundaries and classify the action. The two-stage methods first generate proposals, then refine the boundaries and classify the action. In order to obtain high-quality detection results, the proposals should precisely cover action with high recalls and reliable confidence scores. There are mainly two types of proposal generation methods, ``top-down'' and ``bottom-up''. The ``top-down'' method~\cite{turn,gtan,rapnet,talnet,ssad} obtains the final proposals by refining the boundaries of anchors or sliding windows with pre-defined scales and calculating the confidence. 
The ``bottom-up'' method~\cite{tag,bsn,bmn,bcgnn,bsn++,gtad} generates proposals by calculating the boundary confidence of each position and matching start positions with end positions. 


The above ``bottom-up'' methods generate proposals by the confidence scores of boundaries and matching maps. 
However, there are some difficulties on boundary-level and proposal-level context aggregation with this framework. 
First, on the boundary level, different actions vary at different speeds.
The boundary of a slow action usually is not a clear temporal position but a transitional interval.
So there is not enough semantic information for the precise predictions of these boundaries without effective local temporal context aggregation.
On the other hand, as shown in Figure~\ref{fig:intr}(a), the start and end boundaries of some actions are so similar that the high confidence of start and end at such positions will generate many invalid matchings without long-range temporal context aggregation.
Second, on the proposal level, it is not proper to simply perform context aggregation on the matching map.
As shown in Figure~\ref{fig:intr}(b), aggregating adjacent matchings with different temporal scales and semantic densities will damage the internal semantic representation of the matchings.
Moreover, adjacent matchings are highly overlapped so that their semantic information is too similar to obtain sufficient semantic supplement after aggregation. 
Therefore, it is necessary to design effective context aggregation methods on the temporal and proposal levels.

To mitigate the above issues, we propose a novel method called {\em Dual Context Aggregation Network} (DCAN) for improving temporal proposal generation. 
Similar to BMN~\cite{bmn}, DCAN has a temporal evaluation branch and a matching evaluation branch. 
For the temporal evaluation branch, we design the {\em Multi-Path Temporal Context Aggregation} (MTCA) to achieve effective and smooth context aggregation on the boundary level. 
MTCA is a stack of Multi-Path Temporal Convolutions (MPTC).
In each MPTC, there is a long-range path equipped with a dilated convolution to expand the receptive field and achieve long-range context aggregation and a short-range path with a regular convolution to aggregate short-range context.
In order to alleviate the gridding artifacts of dilated convolution, we adopt a sawtooth wave-like heuristic arrangement for MPTCs to ensure the context of each position can be aggregated smoothly.
MTCA gradually expands the receptive field from the frame to the entire video, thereby effectively aggregating the long-range and short-range contexts.
For the matching evaluation branch, we propose the {\em Coarse-to-fine Matching} (CFM) for effective context aggregation on the proposal level.
CFM first generates a coarse matching map using the Group Sampling strategy, then gradually refines the map from coarse to fine through the refinement network.
The coarse map ensures the distinction of semantic information between sparse matchings, and at the same time, aggregates the context of adjacent matchings without damaging the semantic representation.
During the coarse-to-fine process, the relation between matchings is gradually supplemented and restored.
CFM enhances the expressiveness and robustness of the matching context, and the final matching map contains the relation between the matchings.

We conduct extensive experiments on the THUMOS-14 and ActivityNet v1.3 to demonstrate the effectiveness of our Dual Context Aggregation Network (DCAN). 
In summary, our contributions are as follows:
\begin{itemize}
\item{On the boundary level, we propose the Multi-Path Temporal Context Aggregation to aggregate boundaries context and alleviate the gridding artifacts of dilated convolutions.}
\item{On the proposal level, we design the Coarse-to-fine Matching to generate and refine matching maps from coarse to fine, which enhances the expressiveness and robustness of the matching context.}
\item{The experiments prove the high performance of DCAN, which can achieve better performance than other state-of-the-art methods on ActivityNet v1.3 and THUMOS-14.}

\end{itemize}

\begin{figure*}[ht]
  \centering
  \includegraphics[width=1\textwidth]{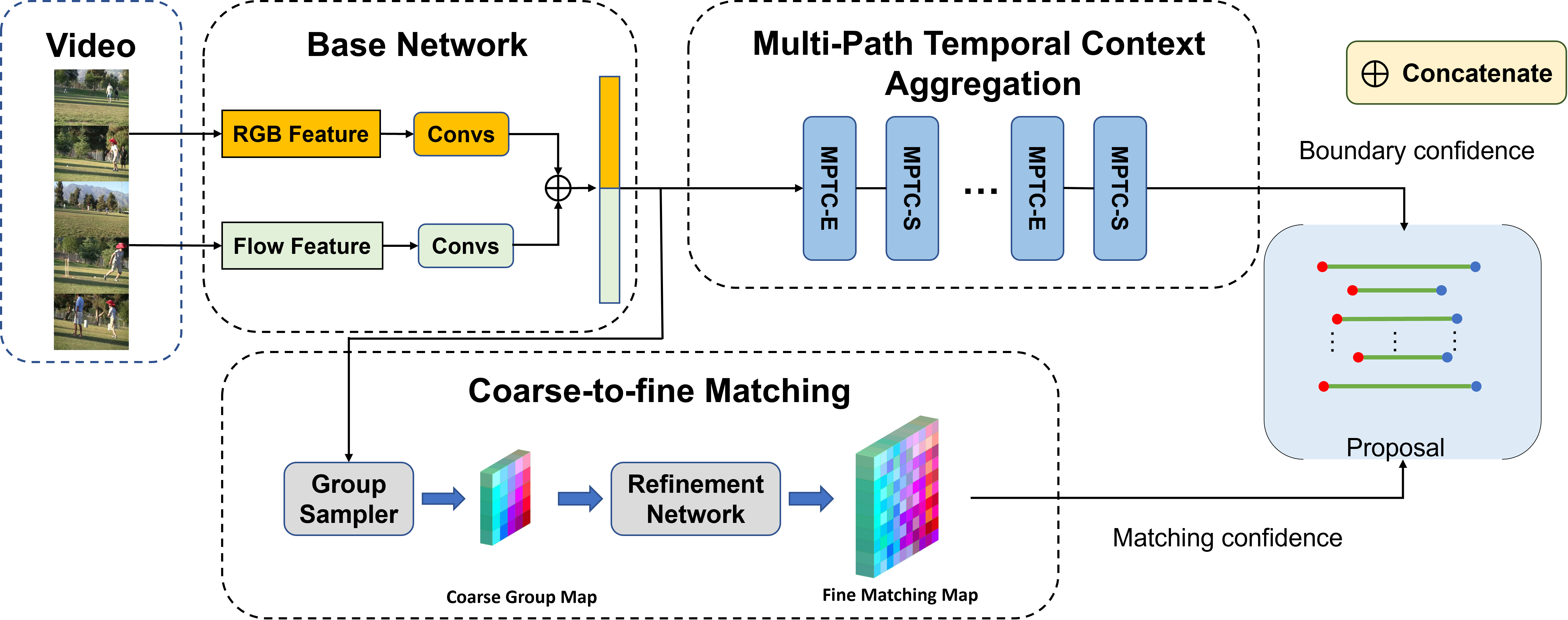}
  \caption{The network architecture of DCAN. First, a dual-path convolution layer is used to model local temporal features with RGB and flow features, respectively. Then, we concatenate these two features and feed them into Multi-Path Temporal Context Aggregation Module to aggregate temporal context for boundary confidence evaluation. At the same time, these features are also input into Coarse-to-fine Matching Module, which generates a coarse group map and then refines the map to a fine matching map through a refinement network for matching confidence evaluation. Finally, the boundary and matching confidence will be combined to obtain final proposals.\\}
  \label{fig:arch}
\end{figure*}

\section{Related Work}



\subsection{Action Recognition}
Action recognition is an essential task in video understanding. It performs spatio-temporal modeling on video frames to recognition actions in the video. 2D CNN methods~\cite{two-stream,tsn,tsm,slowfast,teinet,tea,tdn} take RGB and optical flow as input and perform spatial and temporal modeling, respectively. 3D CNN methods~\cite{c3d,r3d,i3d,t3d} capture spatio-temporal information between frames by performing 3D convolution on stacked video frames. ~\cite{p3d,s3d,r2+1d} model spatio-temporal features by decoupling 2D and 1D convolutions for reducing computing resources. In this work, we use 2-stream~\cite{two-stream} to generate the feature sequence of the video.

\subsection{Temporal Action Detection and Proposals}


Current temporal action detection methods can be divided into one-stage and two-stage methods. The one-stage methods simultaneously generate action proposals and corresponding action labels in a single model. The two-stage methods first generate the proposals, then refine the boundaries and classify actions. ~\cite{turn,rapnet,talnet} generate proposals based on a pre-defined sliding window or anchors and train a classifier to filter anchors.  TURN~\cite{turn} utilizes a sliding window and refines boundaries by concatenating the boundary context and internal context of proposals. RapNet~\cite{rapnet} proposes a relation-aware module to capture long-range context between frames. RTD-Net~\cite{rtd} utilizes the transformer decoder to generate a sparse proposal set directly, effectively omitting post-processing steps. Although many anchor-based methods use multi-scale anchors to increase the diversity, the generated proposals are still not flexible enough to cover actions of varying temporal scales.~\cite{bsn,bmn,bsn++} use a flexible way called boundary matching. They predict each frame's start and end confidence, then match the frames with high start and end confidences to generate the proposals and evaluate their confidence. These methods are more difficult to optimize due to the lack of prior knowledge of the anchor.

\subsection{Temporal Modeling in Temporal Action Detection}
Temporal modeling plays an important role in temporal action detection. ~\cite{daps,e2e} use LSTM to generate action proposals. Compared with LSTM, 1D convolution on temporal dimension shows better performance when modeling the long-range temporal structure of actions. \cite{LeaFVRH17,tsanet} and \cite{bsn++}  utilizes temporal convolution and UNet for temporal relationship modeling, respectively. \cite{tcanet} aggregates local and global temporal context by two types of self-attention modules. We use stacked Multi-Path Temporal Convolution to capture the long-term and short-term dependence of the frames.



\section{Approach}
\subsection{Problem Definition}
For an untrimmed video, we denote it as $U=\{u_t\}^{l_{v}}_{t=1}$, where $l_v$ indicates the length of the video and  $u_t$ is the $t$-th frame. We denote the temporal annotation of action instances as $\Psi_{\rm g}=\{\varphi_n=(t_{\rm s},t_{\rm e})\}^{N_{\rm g}}_{n=1}$ in the video $S_v$ which has  $N_{\rm g}$ instance. $t_{\rm s}$ and $t_{\rm e}$ are the start and end boundaries of the instance $\varphi$ respectively. The action detection model generates prediction proposals that should cover $\Psi_{\rm g}$ with high recall and high temporal overlapping.

\subsection{Overview of DCAN}
As shown in Figure~\ref{fig:arch}, DCAN is composed of three modules: Base Network, Multi-Path Temporal Context Aggregation Module, and Coarse-to-fine Matching Module. 
Firstly, the video frames are fed into the Base Network for local temporal modeling. 
Then the features enter into Multi-Path Temporal Context Aggregation Module and Coarse-to-fine Matching Module to perform the boundary-level and proposal-level context aggregation, respectively. 
The aggregated feature will be used for boundary and matching evaluation, finally generating the proposals. 
We present the technical details of three modules in the following sections.

\subsection{Base Network} 


Following recent proposal generation methods~\cite{bsn,bmn}, we take the temporal features which are extracted using the action recognition backbone network with a fixed-interval sliding window as input.
This feature extracting method only extracts the semantic feature of local temporal sequence frames in an isolated window, resulting in a lack of correlation between adjacent windows, so we design a base network to perform local context aggregation on the temporal feature.
The base network has two convolution paths to model RGB features and optical flow features, respectively.
Each path consists of $N_{\rm base}$ 1D convolution layers, with 128 filters, kernel size of 3, stride of 1, followed by a ReLU activation layer.
Finally, we obtain the feature $F_{\rm base}^{\rm rgb}$ and $F_{\rm base}^{\rm flow}$, then feed them to the following modules.

\subsection{Multi-Path Temporal Context Aggregation} 

\begin{figure}[!t]
  \includegraphics[width=0.5\textwidth]{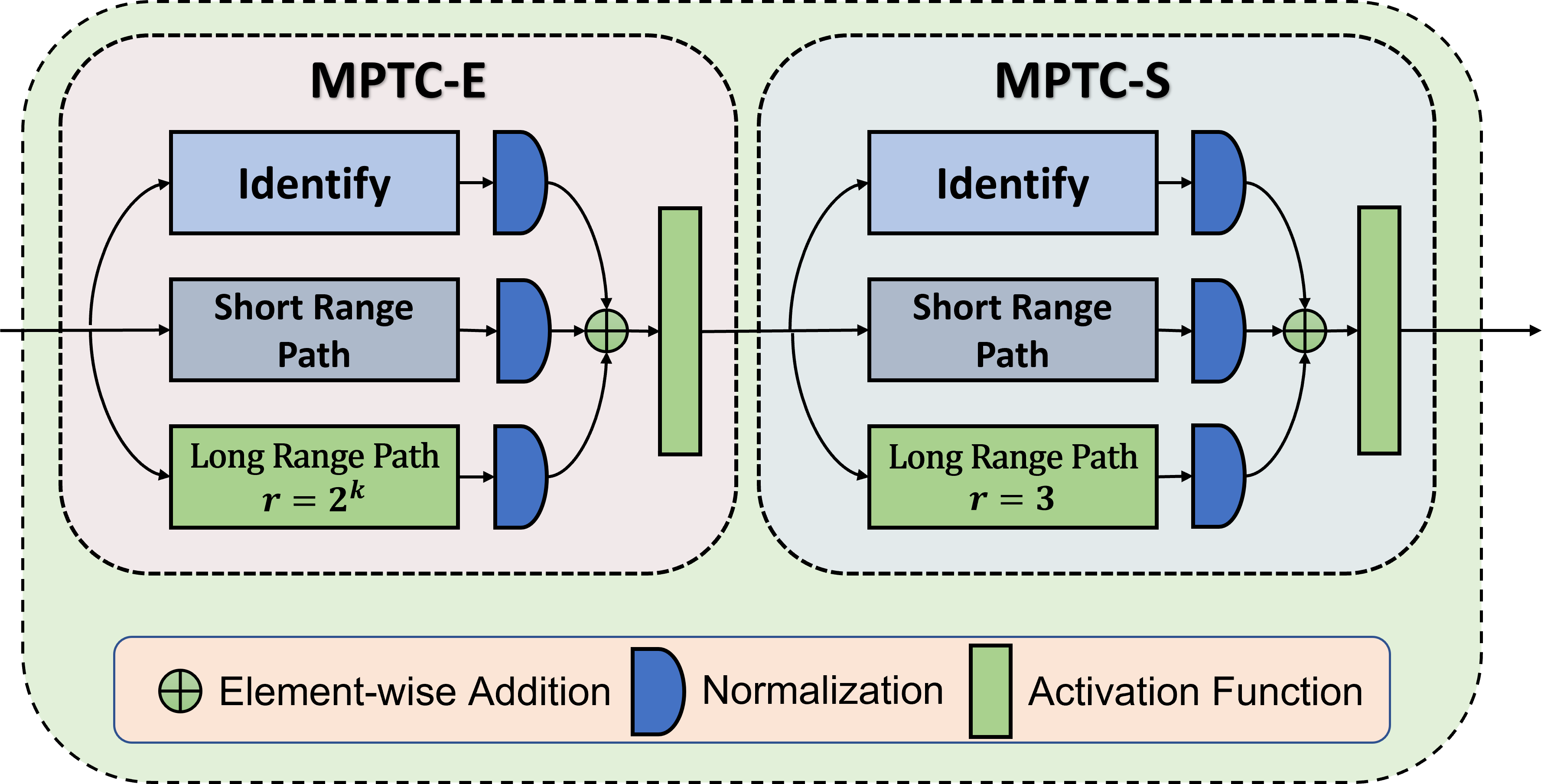}
  \caption{The architecture of MPTC. First, temporal features are fed into MPTC-E with an increasing dilation to expand the receptive field. Then, MPTC-S with a fixed dilation smooths the features from MPTC-E.}
  \label{fig:mptc}
\end{figure}

In this section, we introduce our Multi-Path Temporal Context Aggregation (MTCA) to aggregate long-range and short-range temporal context for temporal evaluation effectively.
As shown in Figure~\ref{fig:arch} and Figure~\ref{fig:mptc}, MTCA is composed of a sequence of stacked Multi-Path Temporal Convolution (MPTC).
In each MPTC, ${\rm Conv_L}$ is the long-range path, including a dilated convolution layer with a kernel size of 3 and a dilation of $r$ to aggregate long-range context and extend the receptive field. 
${\rm Conv_S}$ is the short-range path including a regular convolution with a kernel size of 3 that aggregates short-range temporal context.
In order to enhance the representation ability of features, and at the same time, solve network degradation during training, we introduce a shortcut path to fuse features of different levels.
Finally, we combine these three paths in parallel and do the element-wise addition.
MPTC can be formulated as follows:
\begin{equation}
\label{eqn:mptc}
{\rm MPTC}(x) = \sigma(\varepsilon({\rm Conv_L}(x))+\varepsilon({\rm Conv_S}(x))+ \varepsilon(x)),
\end{equation}
where $\sigma(\cdot)$ and $\varepsilon(\cdot)$ is the nonlinear activation function and the normalization operation respectively.

Previous research~\cite{hdc} on the application of dilated convolution in image detection and segmentation tasks proves that too-rapid expansion of the receptive field may lead to information loss.
Specifically, simply stacking multiple dilated convolution layers with exponentially increased dilation will cause gridding artifacts, that is, features at some positions cannot participate in the calculation.
To alleviate this phenomenon, we design two different types of MPTC. 
The first, called MPTC-E, is an MPTC with dilation of $2^k$ for rapidly expanding the receptive field, where $k$ is the exponent to adjust the scale of receptive field expansion and increases with the depth of the network increases. 
The second, called MPTC-S, is an MPTC with a fixed dilation of $r_{\rm smooth}$ for alleviating the gridding artifacts.

The numbers of MPTC-E and MPTC-S in MTCA are both $N_{\rm b}$, and MPTC-E's dilation $r_i$ is $2^{i}$ where $1\leq i \leq N_{\rm b}$.
To alleviate the gridding artifacts as much as possible, we set the dilations of MPTC-E and MPTC-S are relatively prime and connect these two blocks alternately.
In this way, the top layer of MTCA can access information from the entire video while the aggregation of information at each temporal position is smooth and uniform.
Finally, after aggregating long-range and short-range context at each temporal position, the start and end probabilities are predicted, respectively.



\noindent \textbf{Discussion.}
Some methods use the self-attention mechanism to enhance and aggregate the features of each position, but we argue that the self-attention mechanism is not suitable for boundary evaluation.
The self-attention mechanism pays more attention to the correlation between positions but ignores the order and distance.
For boundary evaluation, context aggregation around the boundary is more valuable than aggregation at a distance.
We hope that the position can focus on aggregating the information of the action instance to which it belongs while also taking into account the aggregation of long-range context rather than the uniform aggregation of the positions of the entire video.
Therefore, MTCA is more beneficial to long-range and short-range context aggregation than the self-attention mechanism.

\subsection{Coarse-to-fine Matching}
Coarse-to-fine Matching (CFM) aims to achieve proposal-level context aggregation by constructing a coarse matching map and refining the map from coarse to fine.
Since adjacent matchings in the matching map have similar sampling intervals and sampling points, which leads lack of distinction and richness in semantics.
Context aggregation of such adjacent matchings cannot obtain an effective semantic supplement.
At the same time, aggregating adjacent matchings with different temporal scales and different semantic densities on the matching map will damage the semantic representation inside each matching.
Hence we propose the Group Sampling strategy to construct the group map.
Specifically, the matchings in the range of $G\times G$ on the original matching map~\cite{bmn} are grouped into a group, and we take the union of their temporal intervals to construct the features of the group.
The entire original matching map can be divided into $\frac{D}{G}\times \frac{T}{G}$ groups without overlap, where $T$ and $D$ is the scale of temporal feature and the maximum duration of any possible action instance.
The strategy can be formulated by:
\begin{equation}
P_{i,j}={\rm GroupSample}(F_{\rm p},s_{i,j},e_{i,j}),
\end{equation}
where ${\rm GroupSample}(F,s,e)$ operation is to uniformly sample $N_{\rm sample}$ points from position $s$ ($0\leq s < e\leq1.0$) to position $e$ of temporal feature $F$. $P_{i,j} \in R^{128 \times N_{\rm sample}}$ is the sampled group feature of the $i$-th row and $j$-th column on the group map. We use hyper-parameter $G$ to set group size and obtain the indices of the group map:
\begin{equation}
0 \leq i < \frac{D}{G},\ 0 \leq j < \frac{T}{G}.
\end{equation}
The corresponding start position $s_{i,j}$ and end position $e_{i,j}$ of sampling and is also formulated by:
\begin{equation}
s_{i,j}=\frac{j\times G}{T},\ e_{i,j}=s_{i,j}+\frac{(i+1)\times G}{T} . 
\end{equation}

Through the above method, we can obtain each group's start and end boundaries and then use the feature construction method of BMN to generate sample-level group map $M_{\rm g} \in R^{\frac{D}{G} \times \frac{T}{G}  \times 128 \times N_{\rm sample}  }$, where 128 is the dimension of $P_{i,j}$. We obtain the group map $M_{\rm g}^{'} \in R^{ \frac{D}{G} \times \frac{T}{G} \times C}$ using a linear transformation to $M_g$.

Then, we refine the coarse group map $M_{\rm g}^{'}$ to the fine matching map $M_{\rm m} \in R^{ D \times T \times C}$ using a refinement network.
The refining process has two steps.
Firstly, we adopt deconvolutions to upsample $M_{\rm g}^{'}$ that each deconvolution layer upsamples the map with factor $2$ on the temporal and duration dimensions, and each group feature is finally refined to $G \times G $ matching features.
Then, a convolution with the kernel size of 3 is adopted to rebuild the relationship of adjacent matchings.
While the refinement network gradually refines group features to matching features, it also restores the relation between matchings and realizes the implicit aggregation of context between matchings.

\noindent\textbf{Discussion.}
The final convolution operation is the same as BMN~\cite{bmn}, but their effects are different.
Our matching features are refined from the group features. 
The construction of the group features weakens the internal temporal scale representation, so it can be considered that the internal temporal semantic representations of different temporal scales matchings are homogeneous.
Performing convolution on such features has a smoother context aggregation effect and can better capture the internal relationship between matchings, so its effect is better than PEM of BMN.



\section{Training and Inference For DCAN}

\subsection{Training}
We train DCAN in the form of a multitask loss function, including a boundary classification loss  $L_{\rm b}$, a proposal evaluation loss $L_{\rm p}$ and a regularization where $\beta$ is set to 0.0001:
\begin{equation}
L=L_{\rm b}+L_{\rm p}+ \beta \cdot L_2(\Theta).
\end{equation}
$L_{\rm b}$ is used to classify whether each frame is the start position or end position of the action:
\begin{equation}
L_{\rm b}=L_{\rm WCE}(P^{\rm start},G^{\rm start})+L_{\rm WCE}(P^{\rm end},G^{\rm end}),
\end{equation}
where $L_{\rm WCE}$ is the weighted binary cross-entropy loss function, $P^{\rm start}$ is the predicted start probability of frames (same for $P^{\rm end}$), $G^{\rm start}$ and $G^{\rm end}$ are the binary ground-truth which are obtained by calculating ${\rm IoR}$ between action instances and temporal positions. The loss $L_{\rm p}$ is used to evaluate proposals confidence. Following BMN, we predict two confidence map $M^{\rm cls}$ and $M^{\rm reg}$, which are trained by the weighted binary cross-entropy loss function and mean squared error loss function respectively:
\begin{equation}
L_p=L_{\rm WCE}(M^{\rm cls},G^{\rm cls})+\lambda \cdot L_{2}(M^{\rm reg},G^{\rm IoU}),
\end{equation}
where $G^{\rm IoU}$ is the IoU map calculated by proposals and ground truth, $G^{\rm cls}$ is the foreground-background map obtained by binarizing $G^{\rm IoU}$ with a threshold 0.9, and $\lambda$ is the loss weight, which is set to 10 as default in our experiments.


\subsection{Inference}
\subsubsection{Score Fusion.}
During the inference stage, the temporal evaluation branch generates the start probability $P^{\rm start}$ and the end probability $P^{\rm end}$ for each position.
We use these two probabilities as the two boundaries scores of the proposals and fuse them with the matching score maps $M^{\rm cls}$ and $M^{\rm reg}$ obtained by the matching evaluation branch to generate the final score of proposals.
Take the proposal  $ \varphi= [t_{\rm s}, t_{\rm e}]$ for example, the combination of final score $p_{\varphi}$ can be shown as:
\begin{equation}
p_{\varphi}= P^{\rm start}_{t_{\rm s}} \cdot P^{\rm end}_{t_{\rm e}} \cdot  (M^{\rm cls}_{t_{\rm e}-t_{\rm s}, t_{\rm s}} \cdot M
^{\rm reg}_{t_{\rm e}-t_{\rm s}, t_{\rm s}})^{\gamma},
\end{equation}
where $\gamma$ is a hyperparameter for adjusting the compatibility of boundary scores and matching scores and is set as 1.5 on THUMOS-14 and 0.8 on ActivityNet v1.3, respectively.

\subsubsection{Post Processing.}
After score fusion, DCAN generates the proposal candidates set as $\Psi_{\rm c}=\{ \varphi_n = (t_{\rm s},t_{\rm e},p) \}^{N_{\rm c}}_{n=1}$ and then we post-process the candidate proposals to remove redundant proposals. 
First, we remove proposals whose start or end probability is lower than half of the corresponding maximum value. 
Then, we adopt the Soft-NMS~\cite{softnms} to eliminate the redundant proposals and obtain the final proposals set $\Psi_{\rm final}=\{ \varphi_n = (t_{\rm s},t_{\rm e},p) \}^{N_{\rm final}}_{n=1}$ , where the number of final proposals to $N_{\rm final}$.

\section{Experiments}
\subsection{Datasets and Setup}

\subsubsection{THUMOS-14.}~\cite{THUMOS14} contains 413 temporal annotated untrimmed videos with 20 action categories. We use 200 videos in the validation set for training and 213 videos in the testing set for evaluation.

\subsubsection{ActivityNet v1.3.}~\cite{anet} is a large-scale action understanding dataset, which consists of 19,994 videos for training, 4,728 for validation, and 5,044 for testing, with 200 action classes. The total duration of the videos is about 600 hours. ActivityNet v1.3 only contains 1.5 occurrences per video on average, and most videos contain a single action category with 36\% background on average.


\subsubsection{Evaluation Metrics.} Average Recall (AR) is the average recall under specified tIoU thresholds for measuring the quality of proposals. On ActivityNet v1.3, these thresholds are set to $\left[0.5:0.05:0.95\right]$. On THUMOS-14, they are set to $\left[0.5:0.05:1.0\right]$. Limiting the average number (AN) of proposals for each video allows us to calculate the area under the AR vs AN curve to obtain AUC. On ActivityNet v1.3, we set AN from 1 to 100. The quality of temporal action detection requires evaluating mean Average Precision(mAP) under multiple tIoU. On ActivityNet v1.3, the tIoU thresholds are set to $\left\{0.5,0.75,0.95\right\}$, and we also test the average mAP of tIoU thresholds between 0.5 and 0.95 with the step of 0.05. On THUMOS-14, these tIoU thresholds are set to $\left\{0.3,0.4,0.5,0.6,0.7\right\}$.

\subsubsection{Implementation Details.}
We use pre-extracted features for all datasets and train the network from scratch. For ActivityNet v1.3, we adopt the two-stream network~\cite{2stream_tsn} fine-tuned on the training set of ActivityNet v1.3 with frame interval $\sigma=16$. Each video feature sequence is rescaled to $L = 100$ snippets using linear interpolation. We set the batch size to 16 and the learning rate to 0.001 for the first 7 epochs and 0.0001 for the following 3 epochs. 
For THUMOS-14, the features are extracted using TSN~\cite{tsn} pre-trained on Kinetics~\cite{kinetics} with $\sigma=5$. We crop each video feature sequence with overlapped windows of size $L = 256$ and stride $128$. In training, we do not use any clips void of actions. We set the batch size to 16 and the learning rate to 0.0004 for all 5 epochs.

The $N_{\rm b}$ is set to 6 on ActivityNet v1.3 and 7 on THUMOS-14. The $N_{base}$, $N_{sample}$, $r_{\rm smooth}$ and $G$ are set to 3, 32, 3 and 2. In the post-processing, the Soft-NMS threshold is set to 0.5 to pick the top $N_{\rm final}$ confident predictions, where $N_{\rm final}$ is 100 for ActivityNet v1.3 and 200 for THUMOS-14.

\begin{table}[t]
\centering
\caption{Comparison of DCAN with other state-of-the-art methods on THUMOS-14 in terms of AR@AN. All models use the two-stream feature as input.}
\label{table:thumos-proposal}
\setlength{\tabcolsep}{1.4 mm}{
\begin{tabular}{c|ccccc}
\toprule
 Method & @50 & @100 & @200 & @500 & @1000 \\
 \midrule
 TAG & 18.55 & 29.00 & 39.61 & - & - \\
 CTAP & 32.49 & 42.61 & 51.97 & - & - \\
 BSN & 37.46 & 46.06 & 53.23 & 61.35 & 65.10 \\
 MGG & 39.93 & 47.75 & 54.65 & 61.36 & 64.06 \\
 BMN & 39.36 & 47.72 & 54.84 & 62.19 & 65.49 \\
 BSN++ & 42.44 & 49.84 & 57.61 & \textbf{65.17} & 66.83 \\
 TCANet & 42.05 & 50.48 & 57.13 & 63.61 & 66.88 \\ 
 \midrule
 \textbf{DCAN} & \textbf{42.65} & \textbf{51.05} & \textbf{57.95} & 64.58 & \textbf{68.37} \\ 
\bottomrule
\end{tabular}
}
\end{table}

\begin{table}[!t]
\centering
\caption{Comparison between DCAN with other state-of-the-art methods on THUMOS-14 dataset. The results are measured by mAP(\%) at different tIoU thresholds. Proposals from all methods are combined with video-level classifier UntrimmedNet~\cite{untrimmednet}.}
\label{table:thumos-map}
\begin{tabular}{c|ccccc}
\toprule
\textbf{Method}  & \textbf{0.7} & \textbf{0.6} & \textbf{0.5} & \textbf{0.4} & \textbf{0.3} \\ 
\midrule
TURN &  6.3 & 14.1 & 24.5 & 35.3 & 46.3 \\
BSN &  20.0 & 28.4 & 36.9 & 45.0 & 53.5 \\
MGG &  21.3 & 29.5 & 37.4 & 46.8 & 53.9 \\
BMN &  20.5 & 29.7 & 38.8 & 47.4 & 56.0 \\
G-TAD &  23.4 & 30.8 & 40.2 & 47.6 & 54.5 \\
BSN++ &  22.8 & 31.9 & 41.3 & 49.5 & 59.9 \\
TCANet &  26.7 & 36.8 & 44.6 & 53.2 & 60.6 \\
\midrule
\textbf{DCAN} &  \textbf{32.6} & \textbf{43.9} & \textbf{54.1} & \textbf{62.7} & \textbf{68.2} \\ 
\bottomrule
\end{tabular}
\end{table}

\begin{table}[t]
\centering
\caption{Comparison with other state-of-the-art methods CTAP~\cite{ctap}, BSN~\cite{bsn}, MGG ~\cite{mgg}, BMN~\cite{bmn} on ActivityNet v1.3 in terms of AR@AN and AUC.}
\label{table:anet1.3-proposal}
\begin{tabular}{c|ccccc}
\toprule
Method & CTAP & BSN & MGG & BMN & DCAN \\ \midrule
AR@1(val) & - & 32.17 & - & - & \textbf{34.42} \\
AR@100(val) & 73.17 & 74.16 & 74.54 & 75.01 & \textbf{75.71} \\
AUC(val) & 65.72 & 66.17 & 66.43 & 67.10 & \textbf{67.93} \\
\bottomrule
\end{tabular}
\end{table}

\begin{table}[!t]
\centering
\caption{Comparison between DCAN with other state-of-the-art methods on ActivityNet v1.3. The results are measured by mAP(\%) at different tIoU thresholds and average mAP(\%). We combined our proposals with video-level classification results from~\cite{cuhk}.}

\label{table:anet1.3-map}
\setlength{\tabcolsep}{3.8 mm}{
\begin{tabular}{c|cccc}
\toprule
\textbf{Method} & \textbf{0.5} & \textbf{0.75} & \textbf{0.95} & \textbf{Average} \\ 
\midrule
SSN & 39.12 & 23.48 & 5.49 & 23.98 \\
BSN & 46.45 & 29.96 & 8.02 & 30.03 \\
BMN & 50.07 & 34.78 & 8.29 & 33.85 \\
G-TAD & 50.36 & 34.60 & 9.02 & 34.09 \\
BC-GNN & 50.56 & 34.75 & 9.37 & 34.26 \\
BSN++ & 51.27 & 35.70 & 8.33 & 34.88 \\
\midrule

\textbf{DCAN} & \textbf{51.78} & \textbf{35.98} & \textbf{9.45} & \textbf{35.39}\\ 

\bottomrule
\end{tabular}
}
\end{table}

\subsection{Comparison with State-of-the-art Results}

\subsubsection{THUMOS-14.} 
We compare DCAN with other state-of-the-art methods on THUMOS-14 in Table~\ref{table:thumos-proposal} and Table~\ref{table:thumos-map}, where DCAN significantly improves the performance of the proposal generation and action detection. 
We report AR@AN for proposal generation and mAP for action detection.
As shown in Table~\ref{table:thumos-proposal}, DCAN improves the AR of all proposals for proposal generation except for @500.
Furthermore, for action detection in Table~\ref{table:thumos-map}, DCAN can also obtain at least 5.0\% improvement when tIoU at any threshold compared to all previous methods.

\subsubsection{ActivityNet v1.3.} 
We compare DCAN with the other methods with the state-of-the-art methods on ActivityNet v1.3 in Table~\ref{table:anet1.3-proposal} and Table~\ref{table:anet1.3-map}. 
We report the AR@AN and AUC for proposal generation and mAP for action detection. 
For a fair comparison, in the proposal generation task, we only compare the methods without the re-sampling strategy.
On two tasks, DCAN outperforms the other state-of-the-art proposal generation methods.
Since DCAN improves AR@1 to 34.42, which demonstrates that the proposals with high confidence also have high recalls, bringing a significant performance improvement in the action detection task.



\subsection{Ablation Studies}


\subsubsection{Ablation comparison with other ``bottom-up'' methods.}
We conduct a direct comparison to other ``bottom-up'' methods, namely, BSN, BMN, and BSN++ in Table~\ref{table:ablation-counterparts} to confirm the effectiveness and superiority of DCAN. 

In the temporal evaluation phase, the TEM of BSN and BMN, which only considers the local details for boundary evaluation, is inferior with limited receptive fields for boundary-level context aggregation. 
BSN++ adopts a shallow U-shaped network to expand the temporal receptive field, but it does not expand the receptive field to global.
MTCA realizes the long-range and short-range smooth context aggregation and expands the receptive field to the entire video.
Therefore, only replacing the TEM with MTCA improves the mAP of ActivityNet v1.3 and THUMOS-14 from 33.85\% and 38.80\% to 35.02\% and 52.55\%.

The BMN and BSN++ directly generate a dense matching map.
This matching feature construction method causes the semantics of adjacent matchings on the map to be similar as well as lack distinction and richness.
In addition, these methods aggregate proposal-level context by applying convolutions or self-attention modules, ignoring that aggregating proposals with multiple different temporal scales and different semantic densities will damage the semantic representation inside the matchings.
CFM refines the coarse group map into the fine matching map. The semantic distinction and richness between adjacent matchings are better, and the temporal scale representation inside the matching is weakened, so the context aggregation is more effective.
After equipping BMN with CFM, the mAPs of the two datasets are increased by 0.95\% and 10.62\%, respectively.

When MTCA and CFM work together, the results of DCAN reach 35.39\% and 54.41\%, which fully demonstrates the importance of context aggregation at two levels and the effectiveness of MTCA and CFM.

\begin{table}[!t]
\centering
\caption{Ablation study results on the validation set of ActivityNet v1.3 and the test set of THUMOS-14. TEB and MEB denote the modules in Temporal Evaluation Branch and Matching Evaluation Branch. We show experimental results on ActivityNet v1.3 in terms of average mAP(\%) and on THUMOS-14 in terms of mAP@0.5(\%).}
\label{table:ablation-counterparts}
\setlength{\tabcolsep}{2 mm}{
\begin{tabular}{c|cc|cc}
\toprule
Model & TEB & MEB  & ANet-1.3 & THU-14 \\ \midrule
BSN & TEM & PEM  & 30.03 & 36.90  \\
BMN & TEM & PEM  & 33.85 & 38.80  \\
BSN++ & CBG & PRB  & 34.88 & 41.30 \\ 
\midrule
DCAN & MTCA & PEM & 35.02 &  52.55  \\
DCAN & TEM & CFM  & 34.80 &  49.42  \\
DCAN & MTCA & CFM  & \textbf{35.39} &  \textbf{54.14} \\
\bottomrule
\end{tabular}
}
\end{table}

\subsubsection{How to choose $r_{\rm smooth}$ of the MPTC-S?}
We conduct experiments to explore how to choose $r_{\rm smooth}$ of MPTC-S for smoothing the receptive field. 
In order to alleviate the gridding artifacts caused by dilated convolution, the dilation of MPTC-S must be mutually prime with the dilation of MPTC-E.
Since the dilation of MPTC-E is set as $2^k$, we choose 1, 3 and 5 as the candidate dilation of MPTC-S.
The experimental results are shown in Table~\ref{table:ablation-dialtion-both}.
We also give the result without MPTC-S and denote it as $*$.
The experimental results show that MPTC-S can improve performance, and the performance is best when $r_{\rm smooth}=3$.

\begin{figure}[!b]
  \includegraphics[width=0.5\textwidth]{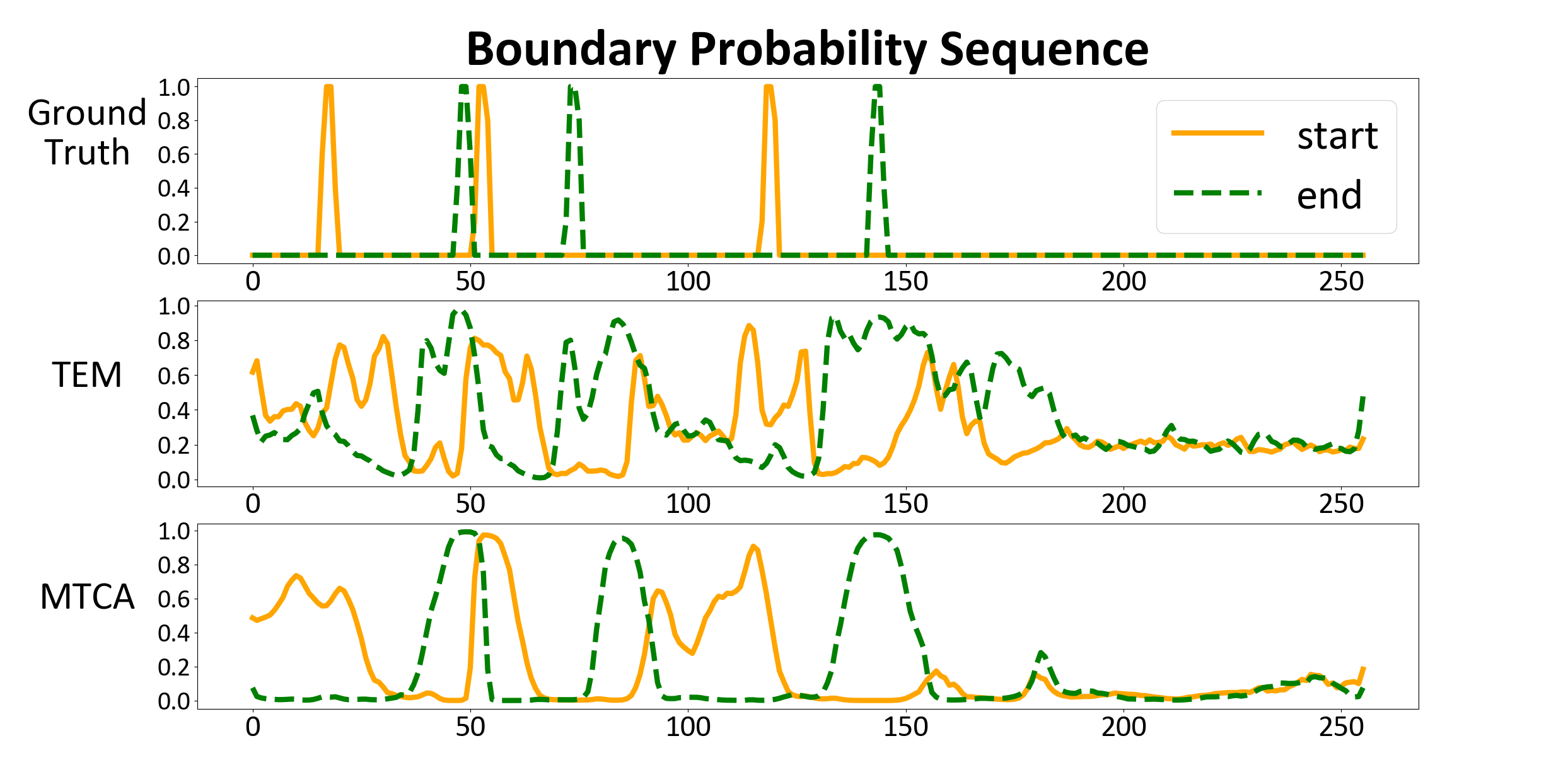}
  \caption{The start and end probability sequences generated by TEM and MTCA.}
  \label{fig:boundary-eval}
\end{figure}

\begin{table}[!t]
\centering
\caption{The effect of different $r_{\rm smooth}$ of the MPTC-S on ActivityNet v1.3 and THUMOS-14 in terms of mAP(\%).}
\label{table:ablation-dialtion-both}
\begin{tabular}{c|c|cccc}
\toprule
Dataset & mAP & $*$ & $r=1$ & $r=3$ & $r=5$ \\
\midrule
 \multirow{4}*{ActivityNet}& 0.5 & 51.49 & 51.56 & 51.78 & \textbf{51.86} \\
 ~& 0.75 & 35.61 & 35.92 & 35.98 & \textbf{36.00} \\
 ~& 0.95 & 8.55 & 8.87 & \textbf{9.45} & 8.86 \\
 ~& Avg & 35.01 & 35.18 & \textbf{35.39} & 35.28 \\
 \midrule
  \multirow{5}*{THUMOS-14}& 0.3 & 30.58 & 32.11 & \textbf{32.62} & 31.48 \\
 ~& 0.4 & 41.40 & 42.17 & \textbf{43.91} & 42.90 \\
 ~& 0.5 & 51.03 & 53.67 & \textbf{54.14} & 53.19 \\
 ~& 0.6 & 59.40 & 61.59 & \textbf{62.73} & 61.36 \\
 ~& 0.7 & 65.16 & 66.80 & \textbf{68.21} & 67.65 \\
\bottomrule
\end{tabular}
\end{table}



\begin{table}[!t]
\centering
\caption{The effect of different group size $G$ of CFM on ActivityNet v1.3 dataset in terms of mAP(\%).}
\label{table:ablation-sparse-rate-anet}
\begin{tabular}{ccccc}
\toprule
$G$ & 0.5 & 0.75 & 0.95 & Average \\
\midrule
1 & 51.34 & 35.82 & 8.90 & 35.11 \\
2 & \textbf{51.78} & \textbf{35.98} & \textbf{9.45} & \textbf{35.39} \\
4 & 51.51 & 35.71 & 8.83 & 35.01 \\
\bottomrule
\end{tabular}
\end{table}



\subsubsection{What is the effect of group size $G$ of CFM?}
We explore the different group size $G$ of CFM, and the experimental results are shown in Table~\ref{table:ablation-sparse-rate-anet}.
When $G=1$, CFM degenerates into the dense matching map similar to BMN.
For $G=2$ and $G=4$, we use one-layer and two-layer refinement networks to refine the map from coarse to fine, respectively.
The experimental results demonstrate that $G=2$ is the best choice.
$G=4$ does not work because when the group range is too large, and the generated map is too coarse, encoding $4 \times 4$ proposals into 1 grouped matchings will lead to the loss of semantic information, and it is difficult to restore this part of the lost information through refinement.

\subsubsection{Generalizability.}

To evaluate the generalizability of proposals, we follow BMN and choose two different subsets on ActivityNet v1.3 for generalization ability analysis. The experiment extracts two un-overlapped action subsets from ActivityNet v1.3: ``Sports, Exercise, and Recreation'' as seen subsets and ``Sports, Exercise, and Recreation'' as unseen subsets separately. Table~\ref{gen} shows the results of DCAN on 2stream features. The results reveal that there is only a slight performance drop on unseen categories, suggesting that DCAN achieves great generalizability to generate high-quality proposals for unseen actions.

\begin{table}[h]
\centering
\caption{Generalizability evaluation on ActivityNet v1.3.}
\label{gen}
\begin{tabular}{ccccc}

\hline
      train            & \multicolumn{2}{c}{Seen(val)} & \multicolumn{2}{c}{Unseen(val)} \\
                  \hline
                  & AR@100         & AUC          & AR@100          & AUC           \\
                  \hline
Seen+Unseen & 74.34          & 66.65        & 75.55           & 67.76        \\
Seen        & 73.43          & 65.10        & \textbf{72.58}           &\textbf{64.86} \\       \hline
\end{tabular}
\end{table}






\subsection{Visualization}
We visualize the start and end probability sequences of BMN's TEM and MTCA in Figure~\ref{fig:boundary-eval}. 
It can be observed that for some boundary positions, TEM is difficult to distinguish whether they are the start or the end boundaries, and some non-boundary positions are recognized as boundaries, which demonstrates only using the local context is difficult to evaluate temporal boundary.
The probability curve of MTCA is smoother and more distinguishable than TEM, and the probabilities of non-boundary position are significantly lower than those of boundaries.
This indicates that long-range and short-range context aggregation on the boundary level can improve the model's ability to distinguish confusing positions and suppress non-boundary positions.

\section{Conclusion}
In this paper, we have proposed a novel Dual Context Aggregation Network (DCAN) for high-quality proposal generation and action detection.
DCAN aggregating context on boundary and proposal level, respectively.
On the boundary level, Multi-Path Temporal Context Aggregation (MTCA) uses multiple paths to aggregate long-range and short-range context smoothly.
Coarse-to-fine Matching (CFM) refines the matching map from coarse to fine and achieves effective context aggregation on the proposal level. 
Extensive experiments on ActivityNet v1.3 and THUMOS-14 demonstrate two-level context aggregation can significantly improve proposal generation and action detection performance.

\section{Acknowledgments}

This work is supported by the National Natural Science Foundation of China (No. 62076119, No. 61921006), Program for Innovative Talents and Entrepreneur in Jiangsu Province, and Collaborative Innovation Center of Novel Software Technology and Industrialization. 

\bibliography{aaai22}

\end{document}